\documentclass[acmlarge, screen]{acmart}

\usepackage{amsmath,amsfonts}
\usepackage{algorithmic}
\usepackage{graphicx}
\usepackage{textcomp}
\usepackage{xcolor}
\usepackage{subcaption}
\usepackage{array} 

\AtBeginDocument{%
  }

\setcopyright{acmlicensed}
\copyrightyear{2025}
\acmYear{2025}
\acmDOI{XXXXXXX.XXXXXXX}
\acmJournal{TIOT}

\newcommand{\dexar}{DeXAR}
\newcommand{\zeroshot}{LLMe2e}
\newcommand{\llmxai}{LLMExplainer}

\begin{document}

\title{Leveraging Large Language Models for Explainable Activity Recognition in Smart Homes: A Critical Evaluation}

\author{Michele Fiori}
\email{michele.fiori@unimi.it}
\orcid{0009-0000-2462-3075}
\author{Gabriele Civitarese}
\orcid{0000-0002-8247-2524}
\email{gabriele.civitarese@unimi.it}
\author{Priyankar Choudhary}
\orcid{0000-0003-2688-7415}
\authornote{Contribution provided while being a postdoctoral fellow at the University of Milan.}
\email{priyankarchoudhary1@gmail.com}
\author{Claudio Bettini}
\orcid{0000-0002-1727-7650}
\email{claudio.bettini@unimi.it}
\affiliation{%
  \institution{EveryWare Lab, Dept. of Computer Science, University of Milan}
  \city{Milan}
  \country{Italy}
}

\renewcommand{\shortauthors}{Fiori et al.}

\begin{abstract}

Explainable Artificial Intelligence (XAI) aims to uncover the inner reasoning of machine learning models. In IoT systems, XAI improves the transparency of models processing sensor data from multiple heterogeneous devices, ensuring end-users understand and trust their outputs. Among the many applications, XAI has also been applied to sensor-based Activities of Daily Living (ADLs) recognition in smart homes. Existing approaches highlight which sensor events are most important for each predicted activity, using simple rules to convert these events into natural language explanations for non-expert users. However, these methods produce rigid explanations lacking natural language flexibility and are not scalable. With the recent rise of Large Language Models (LLMs), it is worth exploring whether they can enhance explanation generation, considering their proven knowledge of human activities. This paper investigates potential approaches to combine XAI and LLMs for sensor-based ADL recognition. We evaluate if LLMs can be used: a) as explainable zero-shot ADL recognition models, avoiding costly labeled data collection, and b) to automate the generation of explanations for existing data-driven XAI approaches when training data is available and the goal is higher recognition rates. Our critical evaluation provides insights into the benefits and challenges of using LLMs for explainable ADL recognition.

\end{abstract}



\received{20 February 2007}
\received[revised]{12 March 2009}
\received[accepted]{5 June 2009}

\maketitle

\textbf{\textit{\textcolor{red}{Paper accepted for publication in the ACM Transactions on Internet Of Things (TIOT) journal.}}}

\section{Introduction}
The Internet of Things (IoT) and pervasive computing communities have investigated sensor-based Human Activity Recognition (HAR) for decades, designing and testing new recognition methods by exploiting the evolution of sensing devices~\cite{babangida2022internet}.
Among the many applications of HAR,
the recognition of Activities of Daily Living (ADL) in smart homes may have high-impact healthcare applications, including the early detection and continuous monitoring of cognitive decline~\cite{bakar2016activity}.

Most existing smart home ADL recognition approaches are based on data-driven deep learning models that require large datasets of (possibly labeled) sensor events. The lack of suitable datasets, especially for the smart home environments, is known as the \emph{data scarcity} problem~\cite{tang2021selfhar}, and together with subject heterogeneity~\cite{soleimani2021cross}, and environment and sensing infrastructure heterogeneity~\cite{chiang2017feature} is one of the reasons of the limited generalization ability of current models.
An additional and not secondary problem is the opacity of 
deep learning models in their activity prediction process.
A few works in recent years proposed the adoption of eXplainable Artificial Intelligence (XAI) techniques to provide the rationale of deep learning ADLs recognition models~\cite{arrotta2022dexar,das2023explainable}.
Several decisions in pervasive computing applications may rely on activity prediction, and inferring \emph{why} an ADL was predicted may lead to more understandable, trusted, and transparent systems~\cite{wolf2019explainability}. This is the case for applications that support clinicians in their diagnosis process~\cite{lussier2018early}.
Existing XAI methods for sensor-based ADL recognition adopt heuristic algorithms to generate sentences in natural language by considering the sensor events in the input time window that were more relevant for the prediction~\cite{das2023explainable,arrotta2022dexar} but the generalization capabilities of those heuristics is unclear. Indeed, these approaches generate rigid explanations that lack the nuance and flexibility of natural language and, at the same time, incorporating new activities or devices would require the significant effort of manually adding new rules.

In this paper, we provide a critical evaluation of the role that LLMs may potentially have in eXplainable sensor-based HAR (XAR).
On the one hand, we investigate how LLMs could be used to associate explanations with activity predictions. On the other hand, we also analyze the risks associated with using LLMs for the generation of explanations. For example, LLMs produce text that appears plausible but can be inaccurate and biased, and it can potentially lead to over-reliance~\cite{hicks2024chatgpt}.

There have been preliminary investigations of applying LLMs to HAR~\cite{cleland2024leveraging}, leveraging the common-sense knowledge about human activities encoded in LLMs to reduce the required labeled data to train supervised classifiers. However, to the best of our knowledge, no existing work studies how LLMs can also be leveraged for explanations and the possible dangers they may introduce.
We aim to answer these research questions: 
\\
Q1) Is it possible to develop a zero-shot ADL recognition method (i.e., not requiring any training data) based on LLM that also provides meaningful explanations? 
\\
Q2) Is it possible to use LLM to automatically generate explanations in natural language, independently from the underlying XAI method adopted?  
\\
Q3) What are the drawbacks and risks of using LLM for XAR?

Our results answer positively to Q1 by implementing an end-to-end LLM-based system and testing it on two public datasets, comparing recognition accuracy with a deep learning baseline and an LLM method that does not provide explanations.
%
We also positively answer to Q2 by proposing a standard format that different XAI methods may adopt to represent the most relevant features leading to the prediction. By using this input and an appropriate prompting technique, LLMs can generate appreciated explanations as proved by user surveys. 
%
%
Regarding Q3, we dedicate the whole Section \ref{sec:discussion} to discuss the potential disadvantages of such systems.
We identify \emph{over-reliance} as the main risk, by providing examples of intuitive LLM explanations for wrong predictions, and propose several mitigation strategies. We also discuss cost and privacy issues, real-life deployment concerns, and the possible impact of LLMs' hallucinations.

To the best of our knowledge, this is the first systematic and critical exploration of using current LLMs capabilities to generate explanations in the Human Activity Recognition (HAR) domain. Moreover, this work proposes a novel pipeline to transform raw sensor data into a structured textual representation for LLMs, also describing how to capture the many sensors that can be deployed in a smart-home environment.

In summary, the main contributions are the following:
\begin{itemize}
    \item We propose \zeroshot{}: a zero-shot LLM method for ADL recognition that also provides human-readable explanations.
    \item We propose \llmxai{}: an LLM-based approach to generate explanations in natural language for ADL recognition in smart homes. \llmxai{} is based on a standardized format to represent the output of XAR methods and, therefore, is agnostic to the specific XAR model being adopted.
    
    \item We conduct extensive experiments showing that \zeroshot{} is slightly less accurate than a state-of-the-art baseline
    while still providing adequate recognition rates. Based on two user surveys with more than $200$ subjects, we show that LLM-based approaches offer explanations that users appreciate more than the ones generated by state-of-the-art heuristic-based methods.
    \item 
    We provide a critical evaluation highlighting the many risks and drawbacks of adopting LLMs in XAR, including examples illustrating the risk of over-reliance on explanations, and proposing several mitigation strategies.
\end{itemize}

\section{Related Work}
\subsection{Explainable Human Activity Recognition}

Several eXplainable Artificial Intelligence (XAI) approaches have been proposed to explain the output of smart home ADL recognition based on deep learning models~\cite{arrotta2022dexar, arrotta2022explaining,das2023explainable}. 
For instance, the work in~\cite{das2023explainable} uses post hoc XAI methods such as LIME~\cite{ribeiro2016should} and SHAP~\cite{scott2017unified} to derive the most important sensor events that an LSTM model used to predict the activity. Post hoc methods consider the deep learning model as a black box, and they analyze the correlations between the classifier's inputs and outputs to determine the most important features for the prediction. On the other hand, the work in~\cite{arrotta2022dexar} adopts a convolutional neural network designed to be explainable~\cite{rudin2019stop}, learning prototypes of input data during training and using them to explain the decision process during classification.
%
In these works, explanations in natural language are generated with heuristic approaches (i.e., simple rules) and describe which sensor events the classifier considers important to perform a prediction. 
%
Such approaches are quite rigid in generating explanations, lacking the nuance and flexibility of natural language (especially with long explanations). At the same time, they lack scalability: incorporating new activities or devices would require manually adding new rules.
To the best of our knowledge, LLMs have never been explored as a flexible method to generate explanations for ADLs recognition.
A major challenge in XAI is also to quantitatively evaluate the effectiveness of the generated explanations~\cite{mohseni2021multidisciplinary}. 
The most common method is to perform user surveys~\cite{das2023explainable,arrotta2022dexar,jeyakumar2023x} and we follow this approach.
An interesting recent work shows that LLMs could also be used to evaluate explanations, and can provide results very close to those of a survey \cite{fiori2024using}.


\subsection{Large Language Models for Human Activity Recognition}

The HAR community is actively working on leveraging Large Language Models (LLMs). Several existing approaches take advantage of the intrinsic knowledge of LLMs about human activities to reduce the amount of required training data, mostly using inertial sensors obtained from mobile and wearable devices~\cite{liu2023large,okita2023towards,zhou2023tent,zilelioglu2023conditional,arrotta2024contextgpt,leng2024imugpt,sharma2025sensorgpt,haresamudram2025limitations,li2024sensorllm,chen2024sensor2text}.
Interestingly, some of them considered the challenging setting of using LLMs for zero-shot recognition of human activities~\cite{ji2024hargpt, hota2024evaluating}.
These works leverage a textual representation of input sensor data to query the LLM for the activity most likely performed by the monitored subject. Since these approaches directly operate on raw inertial sensor data to detect low-level activities, they can not be directly compared with our approach, leveraging different type of data collected in smart home environments to recognize high-level ADLs.

Notably, LLMs have also been proposed for smart home ADL recognition~\cite{thukral2024layout,bouchabou2021using}, with some recent research proposing zero-shot approaches for ADL recognition, specifically designing prompts enabling LLMs to ``reason'' on textual representation of input sensor data to identify the most likely ADLs performed by the inhabitants~\cite{gao2024unsupervised,cleland2024leveraging,xia2023unsupervised,civitarese2024large, chen2024towards,fritsch2024hierarchical}.
However, all of these works focus only on recognizing the performed activities, without providing explanations that non-expert users may leverage to grasp the rationale behind the decision. 


\subsection{Combining Large Language Models and Explainable AI}
There are only a few works that investigated the role of LLMs in improving eXplainable AI. Note that, in this work, we are not interested in the more challenging task of explaining LLMs output~\cite{zhao2024explainability}, but in the use of LLMs to support explainable AI in the HAR tasks.
For instance, the studies in \cite{ali2023huntgpt,mavrepis2024xai} indicate that LLMs could take the output of XAI approaches (i.e., the prediction and the feature importance) and generating natural language explanations. However, to the best of our knowledge, this approach was never studied in the HAR domain. 

%



\section{LLM-based methods for Explainable ADLs Recognition}
In this section, we present two novel approaches exploring how LLMs can be adopted for XAI in sensor-based ADL recognition. In Section~\ref{sec:discussion}, we will also provide a critical evaluation of these methods in subsequent sections to assess their strengths, limitations, and potential implications.

\subsection{The Explainable Human Activity Recognition Problem}
\label{subsec:formalization}

We first formally define 
the XAR problem that we aim to address in this work. Like many studies in this field, we focus on a smart home setting where either a single individual resides or the sensing system is capable of accurately linking each sensor activation to the specific subject responsible for triggering it.

Depending on its type, each sensor generates a stream of \textit{semantic events}. A semantic event is a tuple $\langle e, st, ts \rangle$ where $e$ defines the involved entity (e.g., fridge door), $st$ the status (e.g., open/close), and $ts$ the timestamp at which the event occurred.
For binary sensors, converting raw measurements into semantic events is straightforward. These sensors naturally produce discrete outputs indicating state changes corresponding to specific points in time. For instance, if the magnetic sensor $M_1$ on the fridge triggers the value $1$ at time $t$, we generate a semantic event $\langle \textit{FridgeDoor}, \textit{Opened}, t \rangle$.
On the other hand, sensors generating continuous values (e.g., temperature, humidity, inertial sensors, mmWave sensors) require more advanced processing to extract meaningful events.
Considering continuous environmental sensors (e.g., temperature, humidity), heuristic thresholds can be applied. For example, if at time $t_1$ the humidity  rises above a 
given threshold, a semantic event $\langle \textit{HighHumidity}, \textit{Start}, t_1 \rangle$ is generated. Similarly, when the value falls below the threshold at time $t_2$, the event $\langle \textit{HighHumidity}, \textit{End}, t_2 \rangle$ is generated. Finally, for more complex continuous sensors (e.g., inertial sensors, mmWave sensors, audio sensors), raw data (possibly preprocessed) is fed to simple classifiers in charge of detecting simple context conditions. For instance, a posture detection classifier analyzing inertial sensors from a wearable device may recognize at time $t$ that the subject was sitting, thus generating the semantic event $\langle \textit{Sitting}, \textit{Start}, t\rangle$. The resulting multivariate time series corresponding to the whole set of sensors is segmented in temporal windows.

%
%


Given a window \( \mathbf{w} \) in the time series of semantic events,
and a set of candidate activities \( \mathcal{A} = \{A_1, A_2, \dots, A_n\}\), the goal of HAR is to predict the most likely activity $\hat{A}\in\mathcal{A}$ that the subject is performing during the time of that window. This task is usually performed by a classifier \( h \) such that $h\left(\mathbf{w}\right)=\hat{A}$.
Explainable classifiers generally associate a value to each event in $\mathbf{w}$, denoting its importance for the prediction $h\left(\mathbf{w}\right)$. 
%
For each prediction $\hat{A}=h\left(\mathbf{w}\right)$, an explanation $\mathbf{e}(h\left(\mathbf{w}\right))$ is
 defined as the set of events in $\mathbf{w}$ considered by $h$ as the most important for its prediction (e.g., based on a threshold on the importance values). 

For the sake of interpretability, the sequence of events $\mathbf{e}(h(\mathbf{w}))$ is usually reported in various human-readable forms including graphics or natural language, making explicit the type of events, their temporal relationships.





\subsection{\zeroshot{}: Zero-Shot Explainable ADL Recognition}

\begin{figure}[h!]
    \centering
    \includegraphics[width=0.8\textwidth]{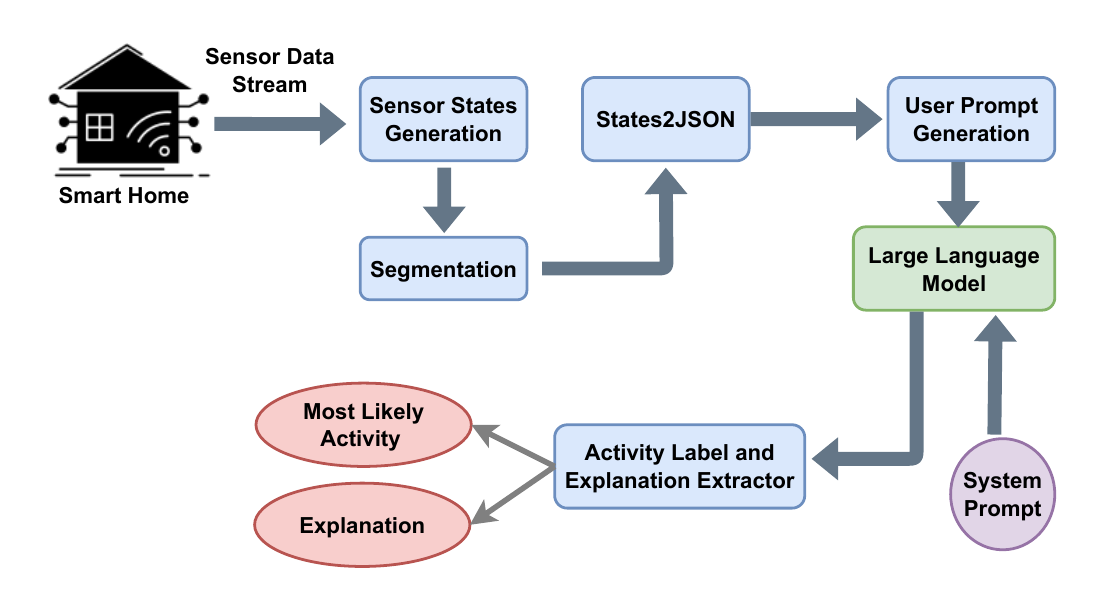}
    \caption{\zeroshot{}: Zero-shot Explainable Activities of Daily Living Recognition}
    \label{fig:zero-shot}
\end{figure}

Starting from the work initially proposed in~\cite{civitarese2024large}, we designed a novel Zero-Shot Explainable ADL recognition method based on LLMs. We will refer to this approach as \zeroshot{}. This method adopts an end-to-end approach, starting with sensor data and using a single LLM prompt to perform both ADL classification and natural language explanation generation simultaneously. \zeroshot{} does not require any training data, since it only leverages the intrinsic knowledge about human activities encoded in LLMs.
The main differences with the work in~\cite{civitarese2024large} are a different representation of the input sensor data and the generation of explanations for each prediction. Note that even other zero-shot approaches proposed in the literature (e.g., ~\cite{cleland2024leveraging}) could be similarly extended to generate explanations.
%
The high-level architecture of \zeroshot{} is shown in Figure~\ref{fig:zero-shot}. 

\begin{figure}[h!]
    \centering
    \includegraphics[width=0.7\textwidth]{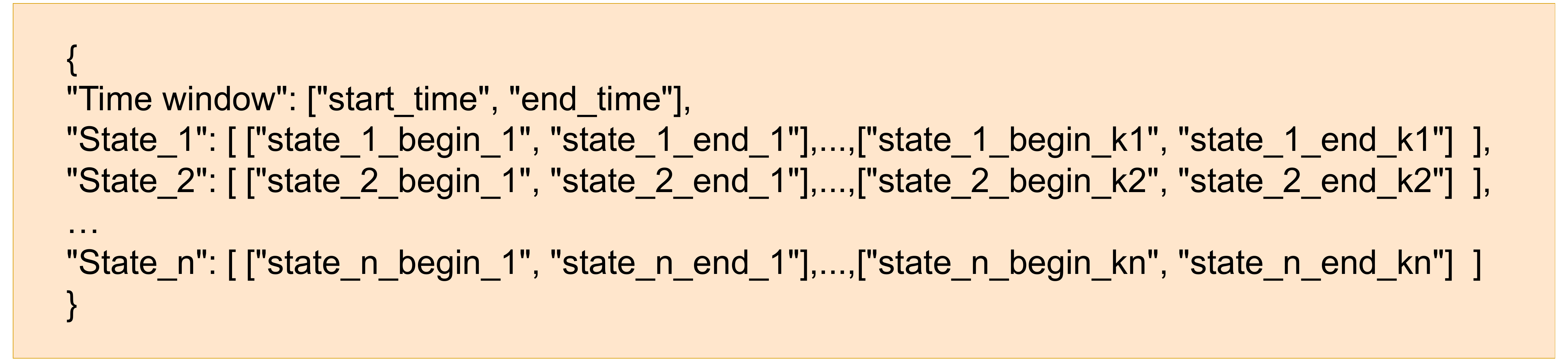}
    \caption{JSON standard representation of the sensor states.}
    \label{fig:jsonformat}
\end{figure}

First, the stream of \textit{semantic events (see Section~\ref{subsec:formalization}) is transformed into a stream of  \textit{sensor states} by the \textsc{Sensor States Generation} module. Intuitively, each sensor state represents the time interval during which a specific property detected by a sensor (e.g., the front door being opened) is true.}
Next, the stream of sensor states is divided into fixed-time windows by the \textsc{Segmentation} module. Each window is then converted into a structured text format by the \textsc{States2JSON} module. 
Figure \ref{fig:jsonformat} shows a JSON standard format that we designed for sensor state representation. This format is used to provide the input to \zeroshot{} for activity prediction, but is also adopted by the \llmxai{} approach to represent the states corresponding to important events identified by XAR techniques, as we will discuss in Section~\ref{subsec:llmxai}.

This textual data is used by the \textsc{User Prompt Generation} module to create the input for the large language model (LLM). The LLM is then queried with a system prompt (i.e., a detailed description of the ADL recognition and explanation generation tasks), along with the user prompt. Finally, the LLM's output is parsed by the \textsc{Activity Label and Explanation Extractor}, which extracts the most likely performed ADL and provides the corresponding explanation.

\subsubsection{System Prompt}

A system prompt has the role of instructing the LLM on the specific task(s) that it has to perform. 
%
The \zeroshot{}'s system prompt adopts a \textit{``role prompting''} strategy, instructing the LLM to operate as a human activity recognition system. This strategy showed to be particularly effective for zero-shot tasks~\cite{kong2023better}. The system prompt includes all the general information that apply to all the windows, in particular: a) information about the layout of the smart home and about the interactions between the subject and the home as they can be captured by the sensing infrastructure; b) the JSON input format and how to interpret it; c) the ADL classification task; d) the explanation generation task. 
%
%

\begin{figure}[h!]
    \centering
    \includegraphics[width=0.7\textwidth]{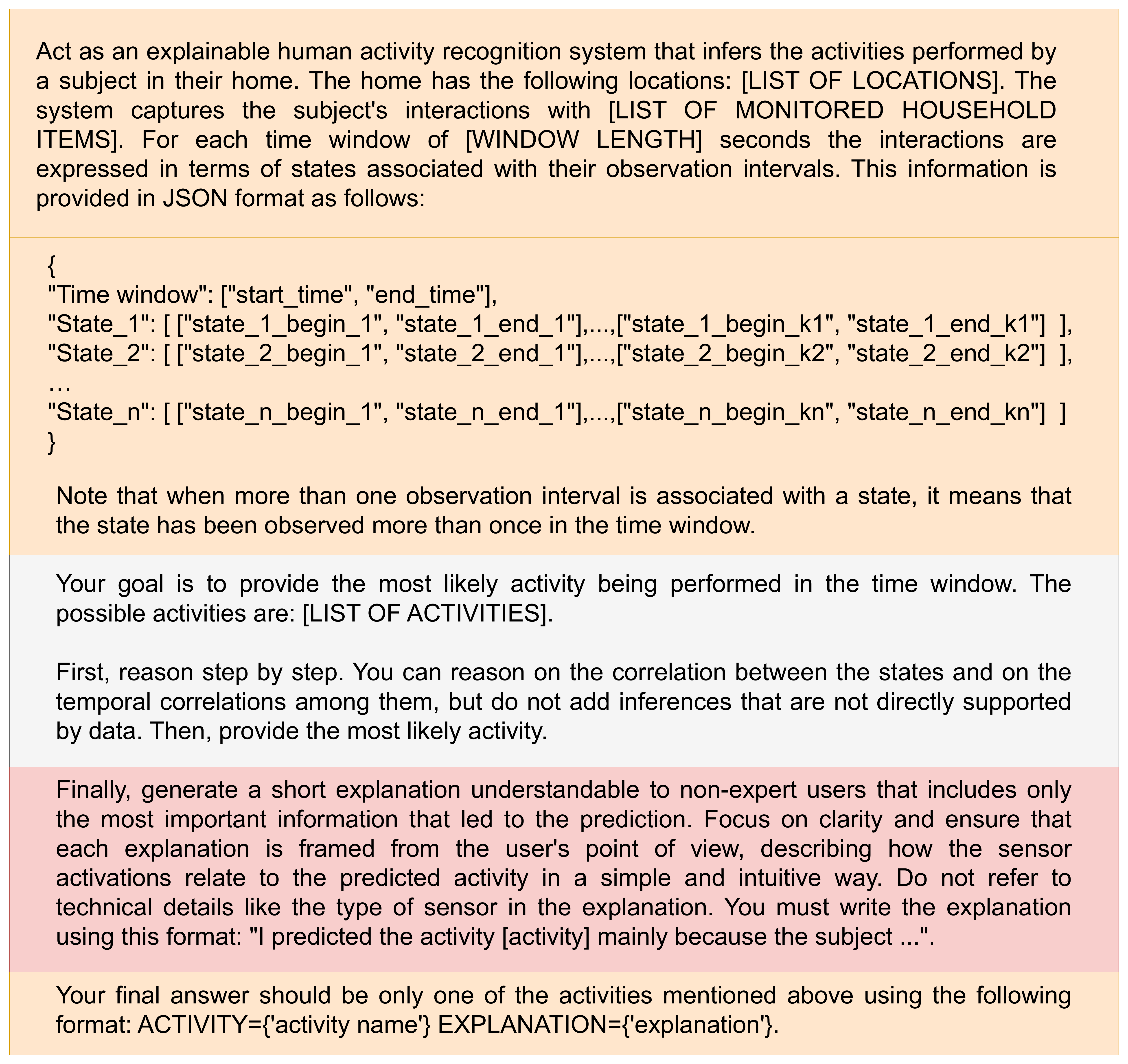}
    \caption{\zeroshot{}: template of the system prompt.}
    \label{fig:llme2e_system_prompt}
\end{figure}

\subsubsection{User Prompt}
\label{subsubsec:s2j}

The user prompt defines a specific instance of the problem that the LLM has to process. In the following, we describe how \zeroshot{} creates each instance from the stream of semantic events introduced in Section~\ref{subsec:formalization}.

\paragraph{\textbf{Step 1: Sensor State Generation}}

First, \zeroshot{} converts the stream of semantic events into a stream of \textit{states}. Each state represents a time interval during which a specific property holds continuously (e.g., the fridge door was open from $t_1$ to $t_2$).
Formally, a semantic state is denoted as $\text{st}(\sigma, t_1, t_2)$, indicating that the property $\sigma$ holds from time $t_1$ to $t_2$. For instance, $st(\textit{MedicineDrawerOpen}, \textit{10:18am},\textit{10:19am})$ represents the fact that the medicine drawer was open in the interval $[\textit{10:18am},\textit{10:19am}]$.  As another example, $st(\textit{OnTheCouch}, \textit{7:45pm},\textit{9:12pm})$ represents the fact that the pressure sensor on the couch reveals that the subject was sitting there between \textit{7:45pm} to \textit{9:12pm}.
To derive a state, we pair semantic events that are complementary (e.g., \textit{Open}/\textit{Close}, \textit{On}/\textit{Off}). Two semantic events $\langle e, s, t_1 \rangle$ and \(\langle e, s', t_2 \rangle\) are paired if $s$ and $s'$ are complementary,  $t_1 < t_2$, and there are no other events from $e$ occurring between $t_1$ and $t_2$.
For instance, the state $st(\textit{OnTheCouch}, \textit{7:45pm},\textit{9:12pm})$ is obtained from the semantic events $\langle \textit{OnTheCouch}, \textit{Start}, \textit{7:45pm} \rangle$ and $\langle \textit{OnTheCouch}, \textit{End}, \textit{9:12pm} \rangle$, given that there are no other $OnTheCouch$ events in the interval $[\textit{7:45pm}, \textit{9:12pm}]$.


\paragraph{\textbf{Step 2: Segmentation}}
The \textsc{Segmentation} module segments the stream of sensor states into fixed-length time windows of  $\tau$ seconds, with an overlap factor of $o$ between consecutive windows. Each time window is associated with the sensor states that are active during that period, but only considering the overlapping portions of the original sensor state intervals. Indeed, we want to ensure that each state is entirely included within the time window.

For example, consider a time window $w$ in the interval $[\textit{3:20pm}, \textit{3:40pm}]$. The semantic state $st(\textit{FridgeOpened}, \textit{3:34pm}, \textit{3:35pm})$ would be entirely contained in $w$. On the other hand, consider the state $st(\textit{TelevisionOn}, \textit{3:12pm}, \textit{3:25pm})$. In this case, while the state begins before the beginning of $w$, we would include only the overlapping portion $st(TelevisionOn, 3:20pm, 3:35pm)$. This step is crucial because we want to provide the LLM only with the properties that hold during the time window.
%
%
More formally, a semantic state $st(\sigma, t_1, t_2)$ is considered for inclusion in a window $w$ if $[t_1, t_2] \cap [t_1^w, t_2^w] \neq \emptyset$ with $[t_1^w, t_2^w$] being the interval of $w$ (i.e., the state has some time instants in common with $w$).
The actual state that is included is $st(\sigma, t_1', t_2' )$
where $t_1' = \max(t_1, t_1^w)$ and $t_2' = \min(t_2, t_2^w)$.


\paragraph{\textbf{Step 3: States2Json}}
For each window $w$ of semantic states, the \textsc{States2Json} module creates a JSON representation, structured as follows.
The \textit{``Time window''} field is represented as an array of two timestamps, formatted as "HH:MM:SS", marking the start and end of $w$. This field is crucial to provide temporal context to the LLM.
Then, for each unique property $\sigma$ of the semantic states in $w$, the JSON contains a dedicated field named with a ``human readable'' label for the property. For each field, the value is a list of one or more time intervals, one for each occurrence of a state with that property in the window.

\begin{figure}[h!]
    \centering
    \includegraphics[width=0.7\textwidth]{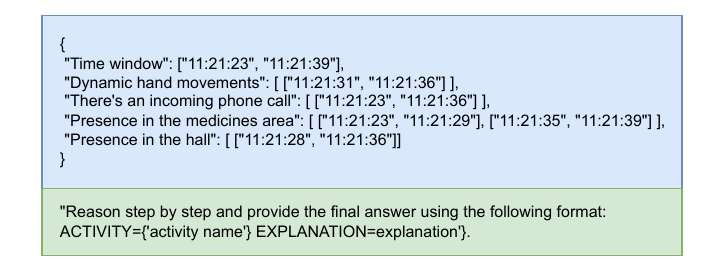}
    \caption{\zeroshot{}: An example of user prompt.}
    \label{fig:user_prompt_example}
\end{figure}

\paragraph{\textbf{Step 4: User Prompt Generation}}
Finally, the user prompt is generated by combining the JSON representation of the input window created by the \textsc{States2JSON} module with instructions to adopt the well-known Chain of Thought (CoT) prompting strategy~\cite{wei2022chain}.  An example of generated user prompt is shown in Figure~\ref{fig:user_prompt_example}.

\begin{figure}[h!]
    \centering
    \includegraphics[width=0.7\textwidth]{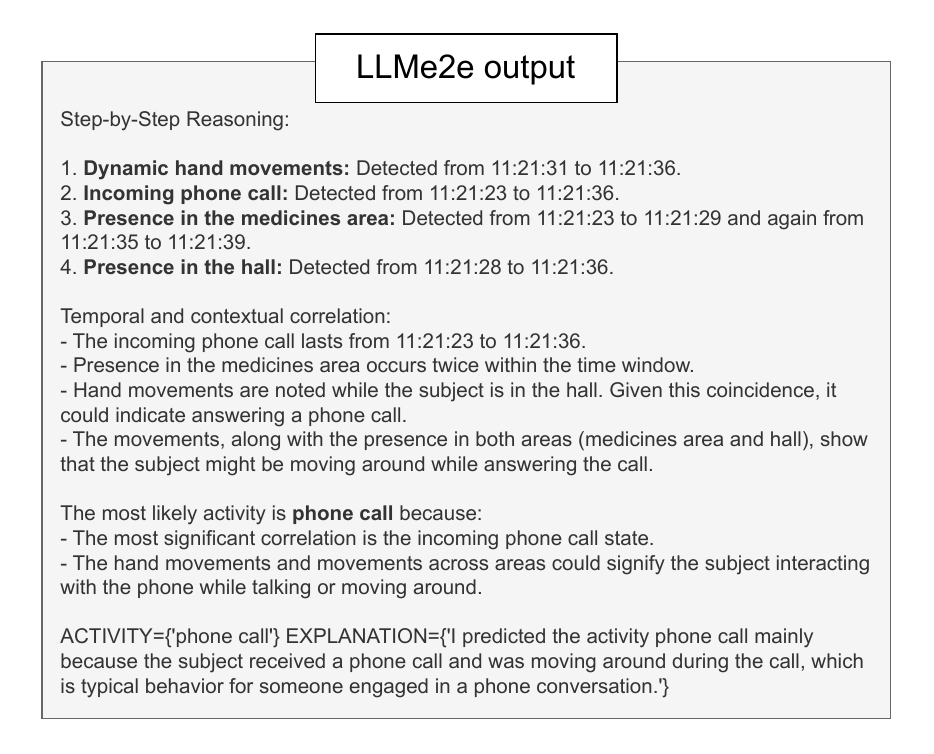}
    \caption{\zeroshot{}: output example}
    \label{fig:llme2e_output_example}
\end{figure}

Figure~\ref{fig:llme2e_output_example} depicts an example of output generated by \zeroshot{} based on the system and user prompts illustrated above.

\subsection{\llmxai{}: LLMs as explanation generators for data-driven XAR}
\label{subsec:llmxai}
\begin{figure}[h!]
    \centering
    \includegraphics[width=0.8\textwidth]{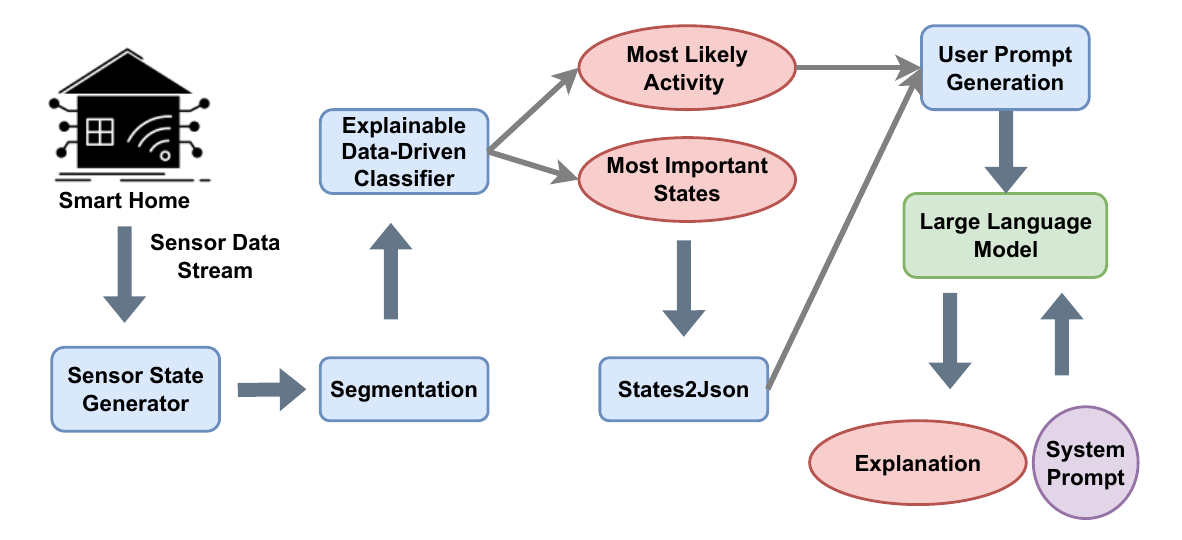}
    \caption{\llmxai{}: Overall architecture.}
    \label{fig:llm-xai}
\end{figure}

In this Section, we propose \llmxai{}: an LLM-based method to generate natural language explanations from the most important events $\mathbf{e}$
derived by a data-driven XAR methods. \llmxai{} is agnostic to the underlying XAR methods being adopted, since the only requirement is that their output, i.e., the most important features contributing to each prediction, is represented as the most important semantic states contributing to classification.
Indeed, \llmxai{} leverages the \textsc{States2Json} model described in Section~\ref{subsubsec:s2j} to convert semantic states into a JSON representation for the LLM.
Considering the existing XAR methods, the conversion in this format is straightforward.
The architecture of \llmxai{} is depicted in Figure~\ref{fig:llm-xai}. 

\subsubsection{System Prompt}
\label{subsec:llmexplainer_system_prompt}
\begin{figure}[h!]
    \centering
    \includegraphics[width=0.7\textwidth]{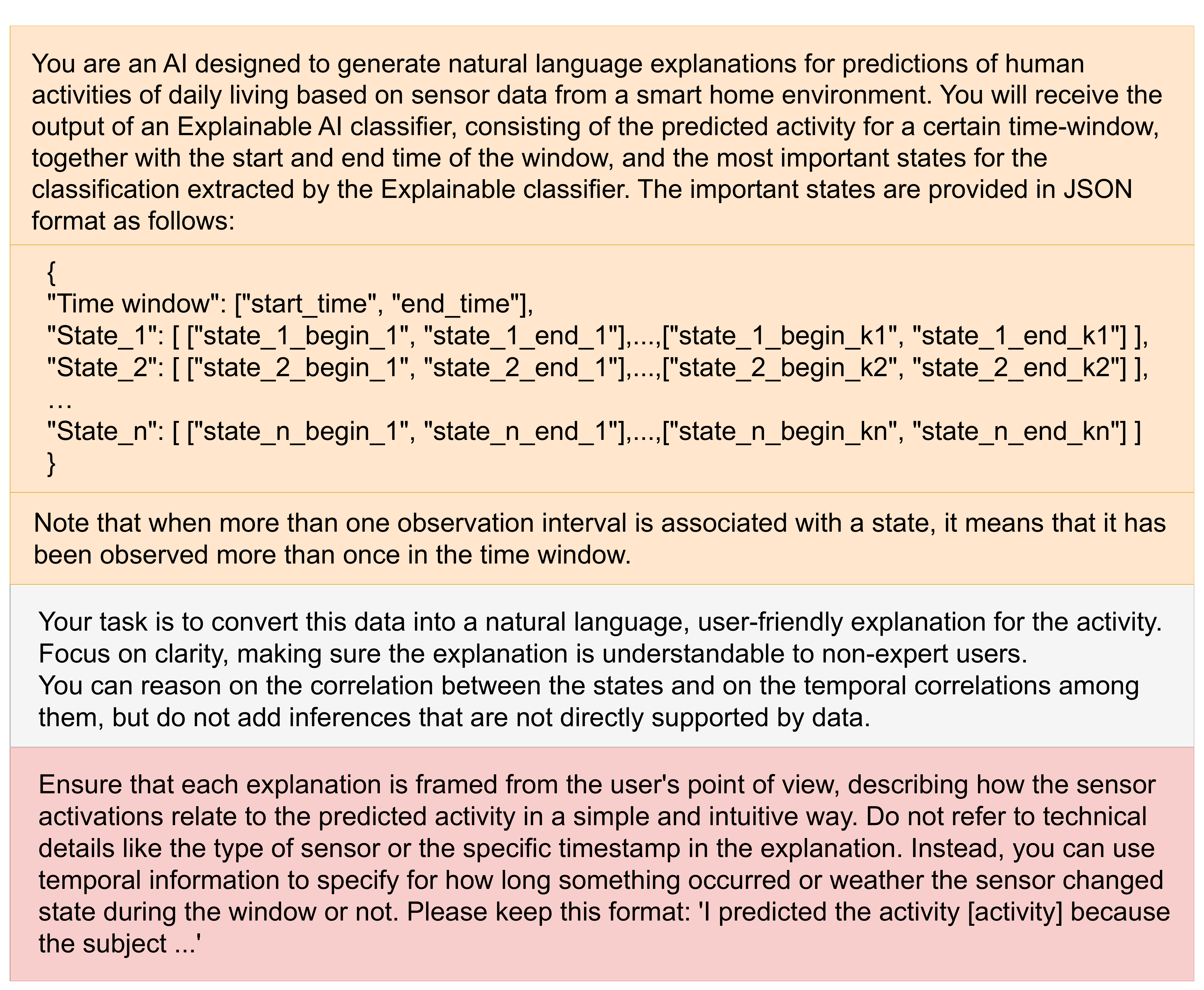}
    \caption{\llmxai{}: System prompt.}
    \label{fig:llm_explainer_system_prompt}
\end{figure}

While the system prompt of \llmxai{} (depicted in Figure~\ref{fig:llm_explainer_system_prompt}) resembles the one of \zeroshot{}, it is only focused on the explanation generation task. 
Specifically, it includes a) a high-level description of the task, using the role-prompting strategy; b) the JSON input format representing the most important states identified by the explainable data-driven classifier and how to interpret it; c) instructions on how to generate the explanation; d) instructions to make sure that the explanations are suitable to non-expert users.
Note in particular the sentence "you can reason on the correlation between the states and on the temporal correlations among them, but do not add inferences that are not directly supported by data". We added this part to better guide the model to stick to the facts stated in the user message.

\subsubsection{User Prompt}

\begin{figure}[h!]
    \centering
    \includegraphics[width=0.7\textwidth]{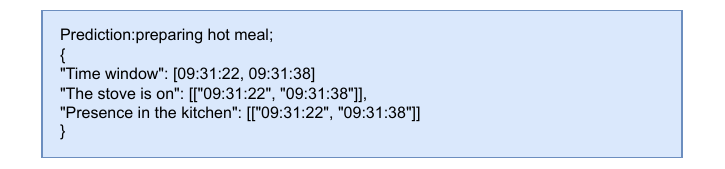}
    \caption{\llmxai{}: Example of an user prompt.}
    \label{fig:llm_explainer_user_message}
\end{figure}

\begin{figure}[h!]
    \centering
    \includegraphics[width=0.7\textwidth]{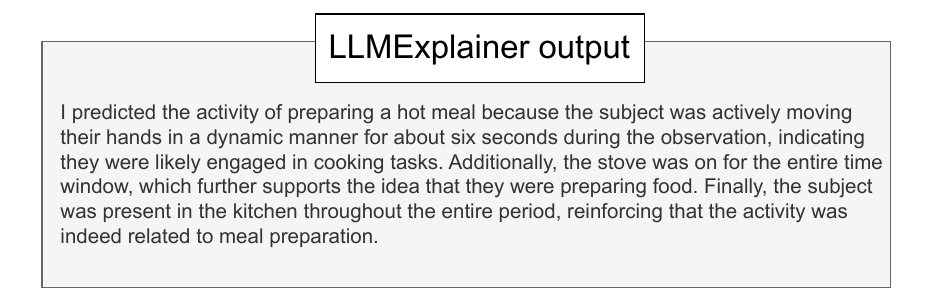}
    \caption{\llmxai{}: Output example }
    \label{fig:llmexplainer_output_example}
\end{figure}

The \textit{User Prompt Generation} module combines the predicted activity and the most important features (formatted using our JSON format) produced by the XAR classifier on a specific window.
An example of a user prompt and the corresponding output are shown in Figures~\ref{fig:llm_explainer_user_message} and~\ref{fig:llmexplainer_output_example}. 
%

\begin{figure}[h!]
    \centering
    \includegraphics[width=0.7\textwidth]{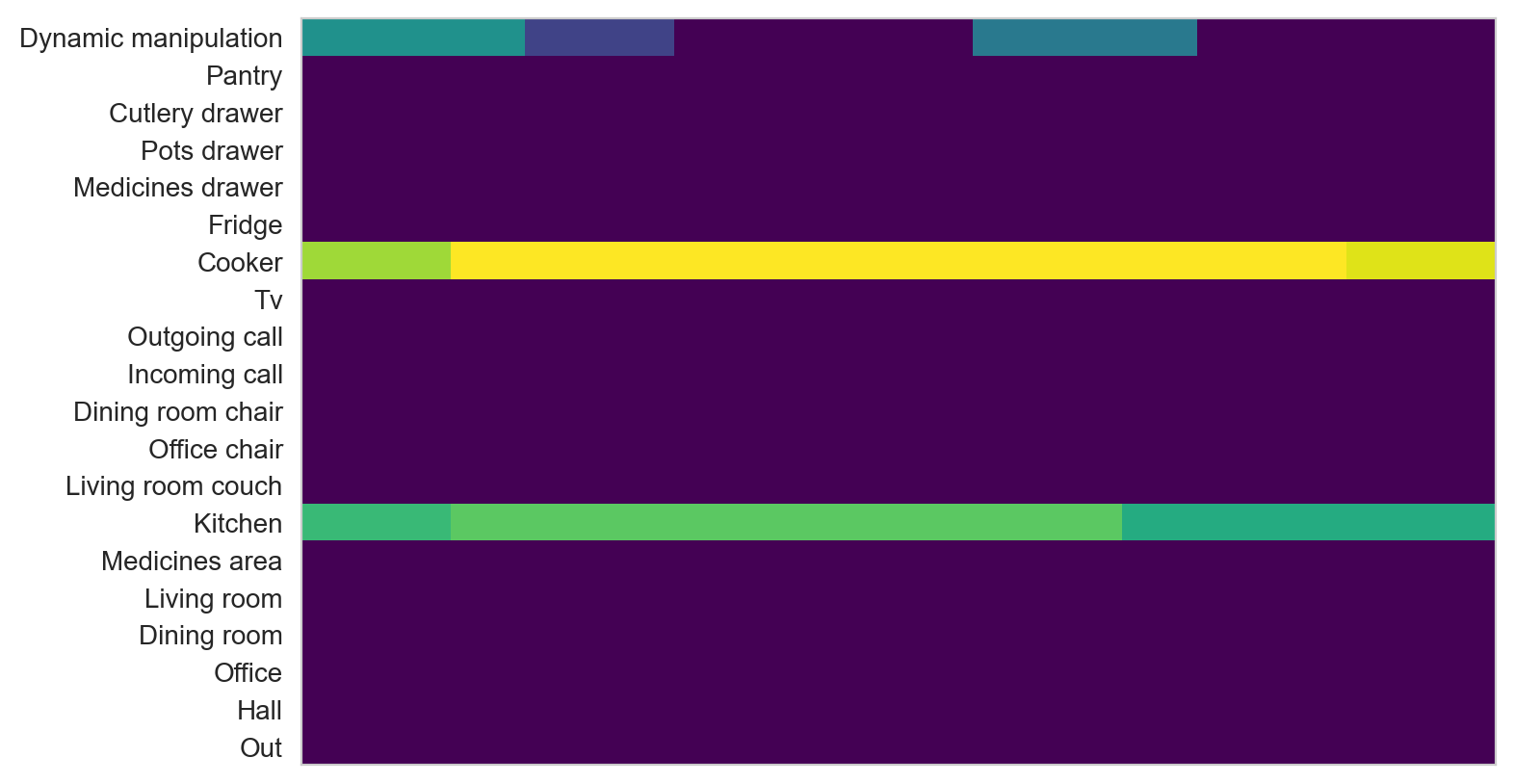}
    \caption{An explanation of  the \dexar{}'s~\cite{arrotta2022dexar} classifier for the predicted activity \textit{preparing hot meal}.}
    \label{fig:dexarsample}
\end{figure}

\begin{figure}[h!]
    \centering
    \includegraphics[width=0.7\columnwidth]{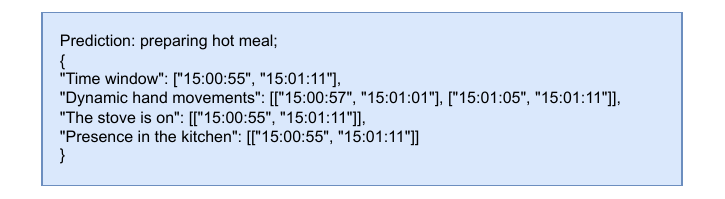}
    \caption{The user prompt for \llmxai{} generated from the most important features in Figure~\ref{fig:dexarsample}.}
    \label{fig:jsondexar}
\end{figure}

As mentioned above, \llmxai{} can be easily used with any XAR model.
However, an additional module converting the most important features derived by a specific XAR model into semantic states is required. 
Indeed, \textsc{States2JSON} expects important features expressed as states. Each state should be associated with a name (e.g., \textit{``the stove is on''}), and the list of intervals in which that state was active.  In order to clarify how \llmxai{} generates the user prompt, we show an example using \dexar{}~\cite{arrotta2022dexar}, representing a state-of-the-art XAR method. The application of \llmxai{} to other XAR approaches like the one presented in~\cite{das2023explainable} is straightforward.
Figure~\ref{fig:dexarsample} shows an explanation (i.e., a heatmap) obtained by \dexar{}'s classifier when predicting the activity \textit{preparing hot meal}.
Intuitively, each row of the heatmap encodes a condition detected by a sensor, and each segment in a row encodes the time interval when the condition was true. In \dexar{}, the most important features are the rows where at least one pixel's intensity is above a threshold.
In order to create the JSON representation, a dedicated module converts each important feature name into a state name and, for each segment, uses its start and end time to represent the time intervals where the state was observed.
Figure~\ref{fig:jsondexar} shows how the most important features from the heatmap in Figure~\ref{fig:dexarsample} would be converted into the JSON format to be used as input for \llmxai{}.

\section{Experimental Evaluation}
In this section, we describe how we evaluated 
\zeroshot{} and \llmxai{} and report experimental results.

\subsection{Datasets}

While validating our approach on complex and real-world datasets would be important, publicly available labeled datasets were mostly collected in scripted settings and/or unrealistic environments and lack real-world diversity.
Hence, for the sake of this work, we selected two publicly available smart-home datasets that include several types of environmental sensors, with one of them including also inertial sensors from a smartwatch.
These two datasets represent different smart home environments and subjects, providing a preliminary indication that our approach generalizes in different conditions.
To the best of our knowledge, the most realistic public and labeled datasets are mostly PIR-based deployments, like CASAS~\cite{casas}. However, due to noise and limited semantics (as we will discuss in detail in Section~\ref{subsec:applicability}), these datasets pose challenges for LLM-based methods and we excluded them from our analysis. Although public and more realistic unlabeled datasets are available, labeled data is essential for our analysis, as it allows us to demonstrate that wrong predictions are often associated with biased explanations leading to over-reliance (as we discuss in Section~\ref{sec:over-reliance}.) 
Hence, the two public datasets that we consider in our experiments are UCI ADL~\cite{ordonez2013activity} and MARBLE~\cite{arrotta2021marble}.


\subsubsection{UCI ADL}
The UCI ADL~\cite{ordonez2013activity} dataset has been collected in two different homes (Home A and Home B), each one inhabited by a single subject. The dataset includes $14$ days of ADLs for Home A and $21$ days for Home B. Both homes are equipped with several types of sensors: movement (PIR), magnetic, pressure, and plug sensors.
Unlike Home A, Home B has PIR sensors also installed over some doors. In contrast, Home A has a PIR sensor above the stove, a plug sensor to detect toaster usage, and a magnetic sensor to monitor interactions with the bathroom cabinet, which are not present in Home B.
In both homes data was collected for the following activities:  \textit{preparing breakfast}, \textit{preparing lunch}, \textit{snacking}, \textit{personal care} (i.e., grooming in the dataset), \textit{showering}, \textit{leaving home}, \textit{relaxing on couch} (i.e., spare time/TV in the dataset), \textit{sleeping}, and \textit{toileting}. Home B includes an additional activity, that is \textit{preparing dinner}. In our experiments, we did not consider the activity \textit{toileting} as the semantic meaning of the label and its relation to the flush sensor is unclear.
Since the \textit{sleeping} and \textit{relaxing on the couch} are over-represented in the dataset (due to a long duration compared to the other ADLs) and exhibit only a few unique sensor patterns, we down-sample these classes in order to have a balanced dataset.


\subsubsection{MARBLE}
The MARBLE~\cite{arrotta2021marble} dataset includes data from $12$ subjects in both single and multi-inhabitant scenarios. For the sake of this work, we only consider the single-inhabitant scenarios of the dataset. The data was collected using smart plugs (to monitor stove and TV usage), magnetic (to monitor the interaction with drawers and cabinets), and pressure sensors (to monitor the usage of dining room chairs and the couch in the living room). Additionally, each subject wears a smartwatch to capture inertial data generated by hand gestures. Consistently with the work in~\cite{arrotta2022dexar}, we only differentiate between \textit{static} and \textit{dynamic} gestures. To detect incoming and outgoing phone calls, an Android application was installed on the smartphone of each subject. 
The original dataset includes 13 activities, but, following \cite{civitarese2024large}, we consider the following $11$ activities: \textit{clearing table}, \textit{eating}, \textit{entering home}, \textit{leaving home}, \textit{phone call}, \textit{preparing cold meal}, \textit{preparing hot meal}, \textit{setting up table}, \textit{taking medicines}, \textit{using pc}, and \textit{watching tv}.

\subsection{Baselines}
In this work, we want to evaluate two aspects: 1) the recognition rate of \zeroshot{}, and 2) the quality of explanations of both \zeroshot{} and \llmxai{}.
We consider the following two baselines:
\begin{itemize}
    \item \textit{\textbf{\dexar{}}}~\cite{arrotta2022dexar}:  a fully supervised approach transforming sensor data into semantic images to leverage Convolutional Neural Networks and XAI methods. Although other baselines could have been considered~\cite{das2023explainable}, this particular one was the only option for which we had access to the code necessary to conduct our experiments.
    We use the \dexar{} implementation leveraging the XAI Model Prototypes approach~\cite{rudin2019stop}, which was considered the best one in the paper. Note that we slightly modified the implementation by removing the extraction of past activities. This is because the approach proposed in~\cite{arrotta2022dexar} can not be applied to \zeroshot{}, since it would require the confidence of the activity prediction that cannot be provided by the LLM.
    In our experiments, we use \dexar{} both as a baseline for explanations and as the XAR method for which \llmxai{} generates explanations. 
    
    \item \textit{\textbf{ADL-LLM}}~\cite{civitarese2024large}: a recent zero-shot ADL recognition method based on LLMs, shown to be superior to other state-of-the-art smart-home LLM-based approaches. We consider this method as a baseline for the recognition rate, as our LLMe2e approach is an extension and adaptation of it. Differently from \zeroshot{}, each window is translated into a sentence in natural language with a heuristic approach, rather than providing just the information of sensors' triggering events like in the JSON template used in \zeroshot{}. 
    ADL-LLM does not explicitly generate explanations, but we consider it to compare the recognition rate.
\end{itemize}

\subsection{Experimental Setup}
We developed the prototypes of \zeroshot{} and \llmxai{} in Python, querying the OpenAI `gpt-4o' LLM model through the official OpenAI library, setting the temperature parameter to $0$. 

\subsubsection{Pre-processing and dataset splitting for ADL recognition}
Considering the segmentation window, we used parameters suggested in the literature. For the MARBLE dataset, we used 16-second windows with 80\% overlap~\cite{arrotta2021marble}, while for the UCI ADL dataset, we used 60-second windows with 80\% overlap~\cite{ordonez2013activity}. Since our approach is zero-shot and does not require training, we could ideally report results using the entire dataset. However, 
since in \dexar{} the dataset was divided into $70\%$ for training and $30\%$ for testing~\cite{arrotta2022dexar}, we evaluate \zeroshot{} using the same test set. The test set for ADL-LLM was $70\%$ and we compare \zeroshot{} on the same set. 


\subsubsection{Quantitative Evaluation}
\label{subsec:user_survey}

The experimental evaluation of our approaches consists of quantitatively measuring both recognition rate and explanation quality.
Considering the recognition rate, we use standard measures like the weighted F1 scores and confusion matrices.

On the other hand, a quantitative assessment of the explanations' quality from the point of view of non-expert subjects requires user-based studies usually adopted in Human Computer Interaction (HCI) works. Indeed, XAI is human-centered, as its primary goal is to make AI decisions understandable and trustworthy to humans. Consequently, evaluating explanations quality inherently requires human judgment
~\cite{mohseni2021multidisciplinary}.

Hence, we performed two user-based surveys: one for the MARBLE dataset and another for the Home B of the UCI ADL dataset. We excluded Home A from this analysis since it is a significantly simpler dataset compared to Home B (as also reflected by our results in Section~\ref{subsec:results}). Frequently, in this dataset there is a only predominant state in the time windows, leading to relatively simplistic explanations.

\begin{figure}[h!]
    \centering
    \includegraphics[width=0.8\linewidth]{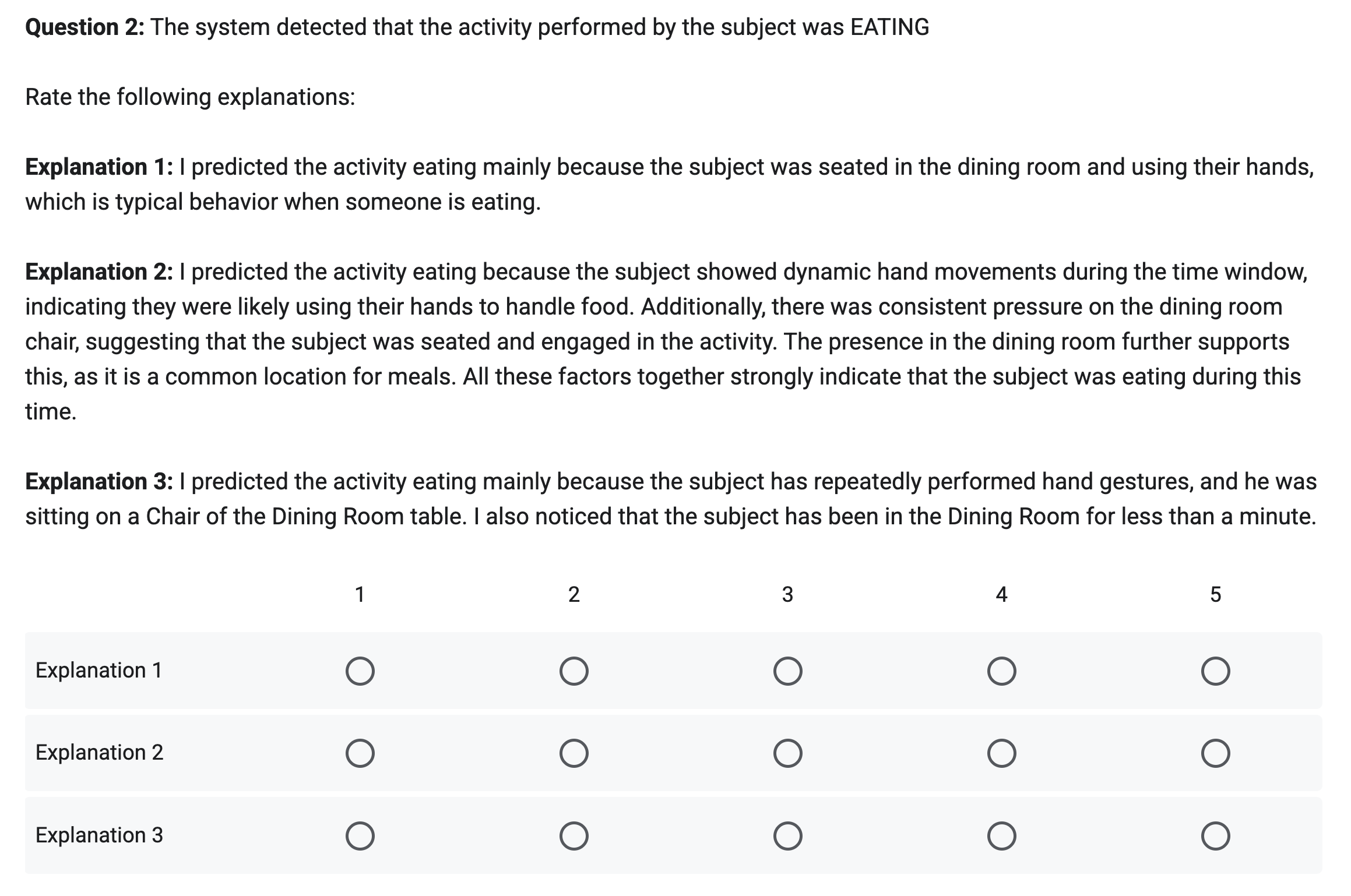}
    \caption{A sample question from our survey}
    \label{fig:surveyscreen}
\end{figure}

Each survey includes, for each activity class, an instance of a window classification associated with the explanations of \dexar{}, \zeroshot{}, and \llmxai{} for that same prediction. Note that in our surveys we included only correct classifications.
In our implementation, \llmxai{} generates explanations using the result of the XAR technique implemented by \dexar{}, as described in Section~\ref{subsec:llmxai}.
Survey participants rated each explanation using the Likert scale. An example of a question in our survey is shown in Figure~\ref{fig:surveyscreen}.

To minimize user subjectivity and ensure statistically significant results, a survey should be administered to a sufficiently large and diverse pool of subjects. For this reason, we took advantage of the Amazon Mechanical Turk platform~\footnote{https://www.mturk.com/} to recruit a total of $366$ unique subjects not having particular experience in HAR.
About one-third of these subjects 
were excluded from the analysis\footnote{Selection was done through attention questions designed to discard bots and users providing random answers.}. Out of the remaining $247$ subjects, $123$ provided answers for the MARBLE dataset, while the remaining $124$ for the UCI dataset. 
Overall, the survey sample consisted of $60\%$ male and 40$\%$ female participants. The majority ($65\%$) were between $31$ and $50$ years old, $30\%$ in the $18$-$30$ age group, and only $4\%$ over the age of $50$. In terms of education, $2\%$ of respondents held a high school diploma, $82\%$ had a Bachelor's degree, $15\%$ possessed a Master's degree, and $1\%$ had achieved a Doctoral level of education.


\subsection{Results}
\label{subsec:results}

\subsubsection{Recognition rate}
Tables~\ref{tab:zs-marble} and ~\ref{tab:zs-uci-dexar} compare the recognition rates of \zeroshot{} (not requiring any training data) and \dexar{} (using a training dataset) measured by the weighted F1-score.

\begin{table}[h!]
\caption{MARBLE: F1 score of \zeroshot{} vs \dexar{} on the same test set (i.e., $30\%$ of the dataset)}
\label{tab:zs-marble}
\centering
\begin{tabular}{|l|c|c|c|c|}
\hline
Activity            & DeXAR  & \zeroshot{}\\ \hline
Clearing table      & 0.36   & 0.37       \\ \hline
Eating              & 0.96   & 0.90       \\ \hline
Entering home       & 0.92   & 0.69       \\ \hline
Leaving home        & 0.90   & 0.80       \\ \hline
Phone call          & 0.98   & 0.82       \\ \hline
Preparing cold meal & 0.59   & 0.54       \\ \hline
Preparing hot meal  & 0.86   & 0.83       \\ \hline
Setting up table    & 0.60   & 0.23       \\ \hline
Taking medicines    & 0.43   & 0.29       \\ \hline
Using pc            & 0.96   & 0.96       \\ \hline
Watching TV         & 0.99   & 0.90       \\ \hline
\hline
\textbf{F1 Weighted avg.} & 0.86 & 0.80 \\ \hline
\end{tabular}
\end{table}

\begin{table}[h!]
\caption{UCI ADL: F1 score of \zeroshot{} vs \dexar{}  on the same test set (i.e., $30\%$ of the dataset)}
\label{tab:zs-uci-dexar}
\centering
\begin{tabular}{|r|c|c|c|c|c|c|c|c|}
\hline
& \multicolumn{2}{c|}{ Home A } & \multicolumn{2}{c|}{ Home B} \\ \hline
\multicolumn{1}{|c|}{Activity} & DeXAR  & \zeroshot{} & DeXAR  & \zeroshot{}  \\  \hline 
Leaving home                   &  1.00  & 1.00      & 0.87   & 0.87  \\ \hline
Personal care                  &  0.99  & 1.00      & 0.93   & 0.95  \\ \hline
Preparing breakfast            &  0.93  & 0.96      & 0.32   & 0.61  \\ \hline
Preparing dinner               &   --   & --        & 0.16   & 0.18  \\ \hline
Preparing lunch                &  0.95  & 0.94      & 0.30   & 0.19  \\ \hline
Relaxing on couch              &  0.99  & 0.99      & 0.88   & 0.76  \\ \hline
Showering                      &  0.99  & 1.00      & 0.91   & 0.93  \\ \hline
Sleeping                       &  1.00  & 1.00      & 0.99   & 0.96  \\ \hline
Snacking                       &  0.00  & 0.40      & 0.31   & 0.51  \\ \hline
\hline
\textbf{F1 Weighted Avg. }        &  0.96  & 0.97      & 0.72   & 0.75  \\ \hline
\end{tabular}
\end{table}

\begin{figure}[h!]
    \centering
    \includegraphics[width=0.6\linewidth]{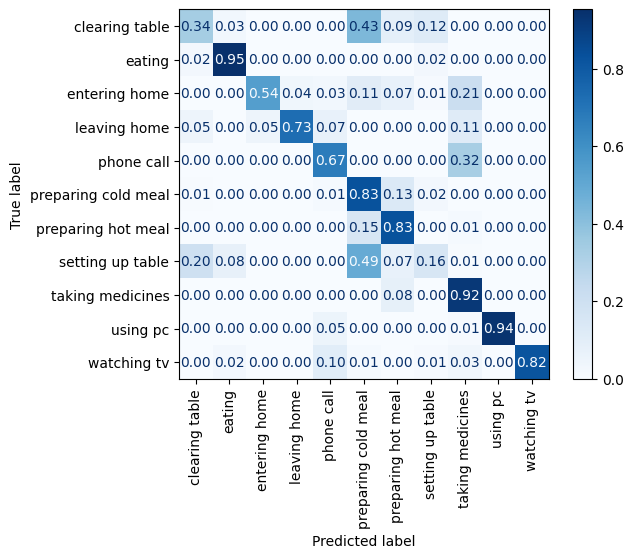}
    \caption{\zeroshot{}: Confusion Matrix on MARBLE dataset}
    \label{fig:cm_marble}
\end{figure}

Considering MARBLE, \dexar{} achieves a weighted F1-score $6\%$ higher than \zeroshot{} indicating the superiority of a method that uses training data. However, \zeroshot{} still reaches an acceptable weighted F1-score of $0.8$.
\zeroshot{} performs significantly worse on the \textit{leaving home}, \textit{entering home}, \textit{setting up table}, and \textit{taking medicines} activities. More details about the recognition errors can be found in the confusion matrix in Figure~\ref{fig:cm_marble}.

In general, the LLM has lower rates for activities that can be learned in a data-driven way by looking at the sensor patterns, while can be easily confused with others by only considering common-sense knowledge.
For instance, \zeroshot{} often misclassifies \textit{setting up the table} as \textit{clearing the table}, \textit{preparing a cold meal}, \textit{preparing a hot meal}, or \textit{eating} due to the similarity between these activities, involving interactions with kitchen items and frequent movement between the kitchen and dining room.

\begin{figure}[h!]
  \centering
  \begin{subfigure}[b]{0.45\linewidth}
    \centering
    \includegraphics[width=\columnwidth]{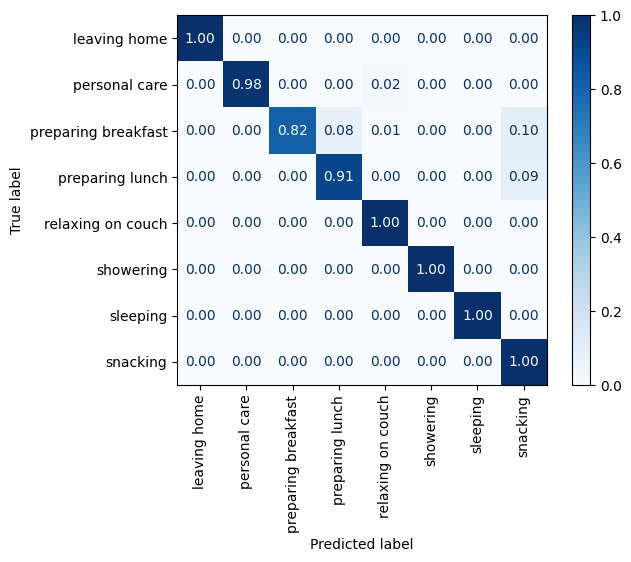}
    \caption{Home A}
  \end{subfigure}
  \hfill
  \begin{subfigure}[b]{0.45\linewidth}
    \centering
    \includegraphics[width=\columnwidth]{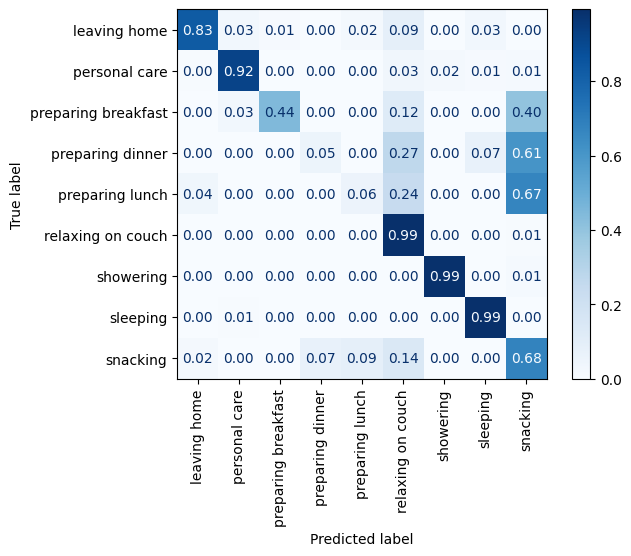}
    \caption{Home B}
  \end{subfigure}
  \caption{\zeroshot{}: Confusion matrices on UCI ADL dataset}
  \label{fig:confusion_matrix_llme2e_uci_adl}
\end{figure}

Considering the UCI ADL dataset, we observe that the recognition rates of \zeroshot{} and  \dexar{} are similar. 
An interesting insight is that, in Home A, \dexar{} can not capture the \textit{Snacking} activity, since it is poorly represented in the training set. On the other hand, even if with a low recognition rate, \zeroshot{} can capture this activity because it does not rely on a training dataset. More details about misclassifications can be found in the confusion matrices in Figure~\ref{fig:confusion_matrix_llme2e_uci_adl}.
For both homes, the main source of misclassification is represented by activities related to eating, since they involve similar sensor patterns.
%
%
%

Tables~\ref{tab:zs-marble_adl} and~\ref{tab:zs-uci-adlllm} compare the recognition rate of ADL-LLM and \zeroshot{} measured by the weighted F1-score\footnote{For the sake of fairness, we did not consider \textit{dynamic manipulations} as input for \zeroshot{} since these were not considered in the ADL-LLM paper~\cite{civitarese2024large}.}. The results show that \zeroshot{} achieves a similar recognition rate for both datasets despite the lower-level input representation (JSON format) and the additional task of generating explanations.

\begin{table}[h!]
\caption{MARBLE: F1 score of \zeroshot{} vs ADL-LLM on the same test set (i.e., $70\%$ of the dataset)}
\label{tab:zs-marble_adl}
\centering
\begin{tabular}{|l|c|c|c|c|}
\hline
Activity            & ADL-LLM & \zeroshot{}  \\ \hline
Clearing table      & 0.25    & 0.14 \\ \hline
Eating              & 0.91    & 0.90 \\ \hline
Entering home       & 0.41    & 0.67 \\ \hline
Leaving home        & 0.71    & 0.81 \\ \hline
Phone call          & 0.97    & 0.82 \\ \hline
Preparing cold meal & 0.51    & 0.52 \\ \hline
Preparing hot meal  & 0.82    & 0.82 \\ \hline
Setting up table    & 0.13    & 0.25 \\ \hline
Taking medicines    & 0.47    & 0.31 \\ \hline
Using pc            & 0.97    & 0.96 \\ \hline
Watching TV         & 0.97    & 0.93 \\ \hline
\hline
\textbf{F1 Weighted avg.} & 0.80 & 0.80 \\ \hline
\end{tabular}
\end{table}


\begin{table}[h!]
\caption{UCI ADL: F1 score of \zeroshot{} vs ADL-LLM on the same test set (i.e., $70\%$ of the dataset)}
\label{tab:zs-uci-adlllm}
\centering
\begin{tabular}{|r|c|c|c|c|c|c|c|c|}
\hline
& \multicolumn{2}{c|}{ Home A } & \multicolumn{2}{c|}{ Home B} \\ \hline
\multicolumn{1}{|c|}{Activity} & ADL-LLM & \zeroshot{} & ADL-LLM & \zeroshot{} \\  \hline 
Leaving home                   & 1.00    & 1.00       & 0.90    & 0.78 \\ \hline
Personal care                  & 0.97    & 0.99       & 0.96    & 0.95 \\ \hline
Preparing breakfast            & 0.75    & 0.90       & 0.71    & 0.75 \\ \hline
Preparing dinner               &  --     & --         & 0.33    & 0.24 \\ \hline
Preparing lunch                & 0.91    & 0.94       & 0.40    & 0.43 \\ \hline
Relaxing on couch              & 0.96    & 0.99       & 0.71    & 0.81 \\ \hline
Showering                      & 0.99    & 1.00       & 0.95    & 0.96 \\ \hline
Sleeping                       & 1.00    & 1.00       & 0.94    & 0.99 \\ \hline
Snacking                       & 0.36    & 0.32       & 0.28    & 0.30 \\ \hline
\hline
\textbf{F1 Weighted Avg. }        & 0.94    & 0.96       & 0.75    & 0.77 \\ \hline
\end{tabular}
\end{table}
\subsubsection{Explanation evaluation}
\begin{figure}[h!]
  \centering
  \begin{subfigure}[b]{0.42\linewidth}
    \centering
    \includegraphics[width=\columnwidth]{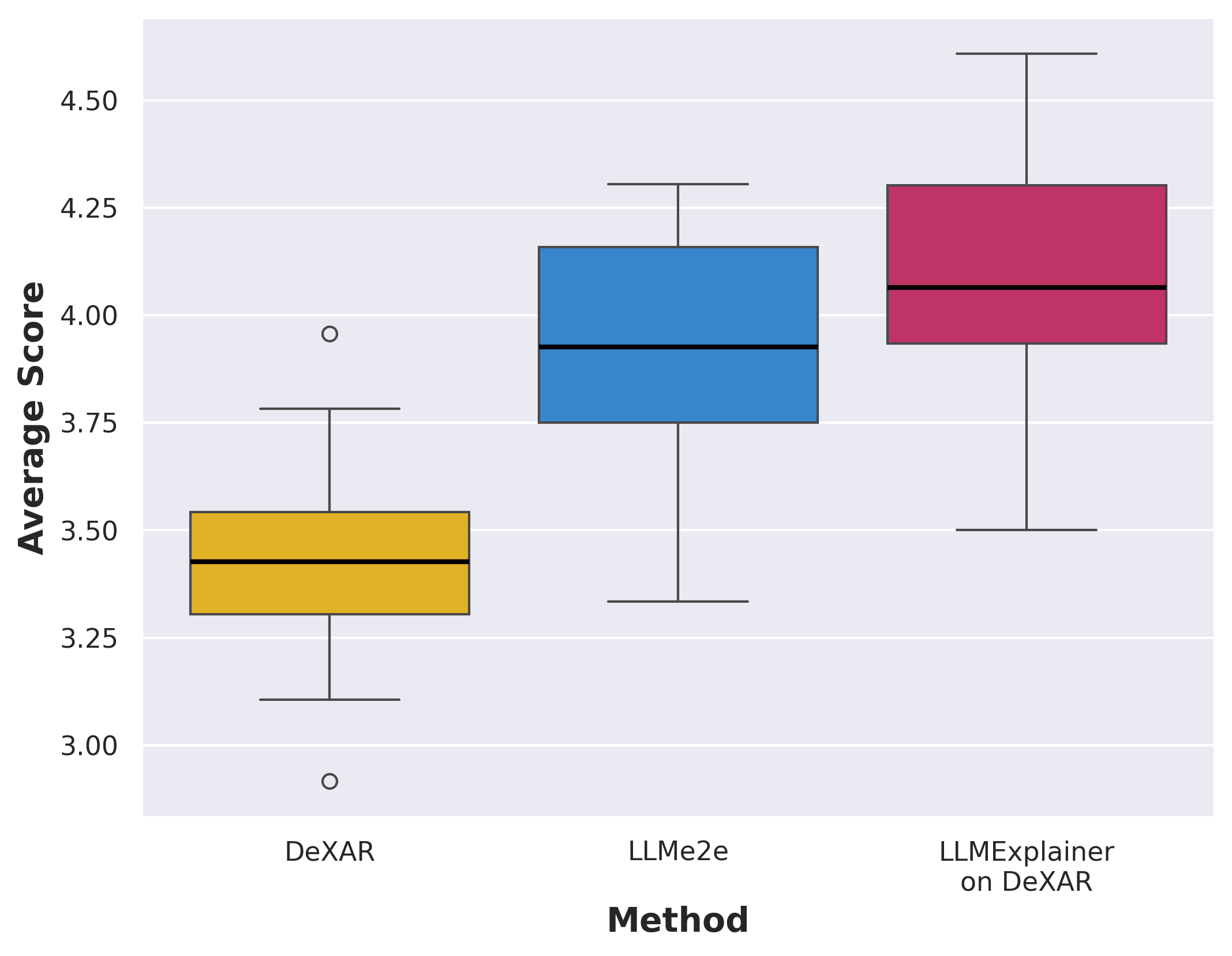}
    \caption{MARBLE}
  \end{subfigure}
  \hfill
  \begin{subfigure}[b]{0.42\linewidth}
    \centering
    \includegraphics[width=\columnwidth]{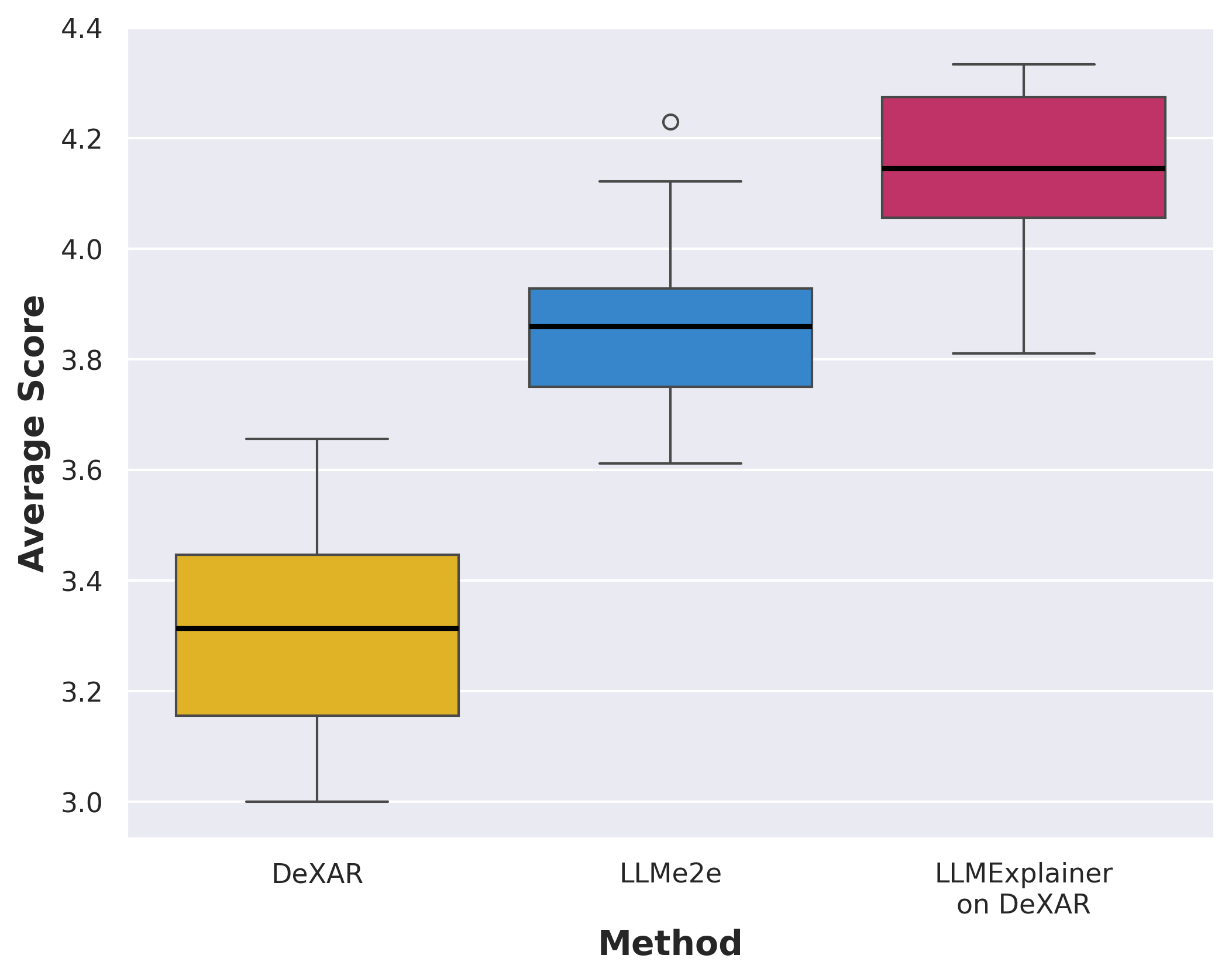}
    \caption{UCI ADL Home B}
  \end{subfigure}
  \caption{Distribution of the scores provided by the participants of the user survey}
  \label{fig:box_plot}
\end{figure}

In the following, we show the results of the user surveys.
The box plots in Figure~\ref{fig:box_plot} depict the distribution of the scores obtained by \dexar{}, \zeroshot{}, and \llmxai{} applied to the output of the \dexar{} classifier.

For both datasets, the participating subjects gave a good score to the explanations generated by the LLM methods with \llmxai{} being the best, while they gave lower scores to the explanations produced by \dexar{}. Even though these two methods leverage the same relevant sensor data (i.e., \llmxai{} generates explanations starting from the important events identified by \dexar{}'s classifier), the LLM uses a more convincing wording including possible relationships between the states detected by sensors and the specific actions done by the inhabitant to perform the predicted activity. 

The scores assigned to the explanations of \zeroshot{} are just slightly lower than the ones assigned to \llmxai{}. 
Overall, these results suggest that when no labeled data is available, \zeroshot{} is a good option since it provides reasonable recognition rates and it offers positively valued explanations. 
On the other hand, when training data is available, a combined technique with \llmxai{} applied to the output of a data-driven XAR method can lead to higher recognition rates and even more appreciated explanations. 
However, in the next section, we will discuss that, even though \llmxai{} explanations were highly appreciated, they may significantly increase the risk of over-reliance.
\section{Drawbacks and Risks of Using LLMs to Generate Explanations}

\label{sec:discussion}

\subsection{Over-reliance}
\label{sec:over-reliance}

Even though XAR may enable non-expert users to better understand the model's predictions and increase their trust, natural language explanations that speculate over the simple facts detected by sensor data may lead to the over-reliance problem. Indeed, a user may excessively trust the system because it generates convincing explanations.
 
While this problem also occurs with existing approaches like \dexar{}, it may be worsened when using LLMs, which are known to produce text that seems plausible but not always necessarily accurate~\cite{hicks2024chatgpt}.


Considering the methods proposed in this paper, \zeroshot{} has complete knowledge of the input data and of the ADL classification process to generate explanations, potentially offering a more accurate explanation of the decision process. On the other hand, \llmxai{} only has to generate an explanation given the most important features obtained from a data-driven classifier. We observed that \llmxai{}, in some instances, 
tends to generate explanations that are linguistically good but not supported by the observed sensor data.
Hence, even though \llmxai{} produces appealing explanations highly rated in our surveys, these can be sometimes misleading, fostering an illusion of certainty and reliability. This problem becomes relevant when considering wrong predictions, as illustrated in the examples of Table~\ref{tab:example_table}.

\begin{table}[h!]
    \centering
    \small
    \caption{Examples of misclassified activity and over-reliance risks}
    \label{tab:example_table}
    \begin{tabular}{|>{\centering\arraybackslash}m{1cm}|>{\centering\arraybackslash}m{3cm}|>{\centering\arraybackslash}m{6cm}|>{\centering\arraybackslash}m{3.2cm}|}
        \hline
        \textbf{Ground Truth} & \textbf{\dexar{}} & \textbf{\llmxai{} on DeXAR}  & \textbf{\zeroshot{}} \\
        \hline
        Watching TV & I predicted the activity \textbf{TAKING MEDICINES} because the subject has repeatedly performed hand gestures and he has been in the dining room. & I predicted the activity of \textbf{TAKING MEDICINES} because the subject showed dynamic hand movements for a short period, which suggests they were likely handling something, such as a medicine bottle or a pill. Additionally, the presence in the dining room during the same time indicates that this activity took place in a common area where people often take their medications. The combination of these factors strongly supports the prediction that the subject was taking their medicines.  & I predicted the activity \textbf{PREPARING COLD MEAL} because the subject moved from the kitchen to the dining room while showing dynamic hand movements, suggesting they were preparing something to eat  \\
        \hline
        Eating & I predicted the activity \textbf{CLEARING TABLE} because the subject has repeatedly performed hand gestures while being sitted in the dining room chair, and he has been in the dining room. & I predicted the activity of \textbf{CLEARING THE TABLE} because the subject showed dynamic hand movements, indicating they were actively using their hands during the task. This activity was accompanied by pressure on the dining room chair, suggesting that the subject was seated while performing the task. Additionally, the presence in the dining room during this time further supports that the subject was engaged in clearing the table. Overall, the combination of these factors points to the activity of clearing the table. &  I predicted the activity \textbf{EATING} mainly because the subject was seated in the dining room and actively using their hands, which is consistent with eating. \\
        \hline
        Preparing hot meal & I predicted the activity \textbf{TAKING MEDICINES} because the subject was sitting on the living room couch while television was on, and he has been in the living room for less than a minute. & I predicted the activity of \textbf{TAKING MEDICINES} because the subject was present in the living room during this time and the living room chair was being used. Additionally, the TV was on, which often indicates a relaxed environment where someone might take their medicines. The combination of these factors suggests that the subject was likely engaged in this activity. & I predicted the activity \textbf{WATCHING TV} mainly because the subject was sitting in the living room with the TV on and was making hand movements, which suggests they were interacting with the TV or a remote control.  \\
        \hline
    \end{tabular}
   
\end{table}

While \dexar{} and \llmxai{} use the same input to generate explanations, \llmxai{} generates significantly longer explanations that aim at convincing the reader that the relevant sensor states obtained by the XAR method are correlated with the predicted activity. 
Despite the system prompts of \zeroshot{} and \llmxai{} contain the same instructions for generating explanations, \zeroshot{} is also aware of the whole classification process. In our experiments this always led to more compact explanations that look less convincing on misclassifications, thus mitigating the over-reliance problem.

Even though it is not possible to completely eliminate this issue, some mitigation strategies can be applied.
The first consists of integrating instructions in the system prompt aimed at avoiding motivations that stick as closely as possible to what is reported in the user message, thus not inventing nonexistent correlations. 
This strategy is already partially 
present in our methodology, in fact, as specified in Section~\ref{subsec:llmexplainer_system_prompt} the system prompt includes the wording "do not add inferences that are not directly supported by data". Nevertheless, we believe that this strategy deserves further investigation and may lead to further mitigation of the problem we are observing.

Another possible and complementary strategy consists of passing to \llmxai{} also the input of the XAR model (i.e. the window of sensor events). In this way the LLM can use this additional information to produce a more factual explanation. Such approach will also produce an LLM-based prediction that could be combined with the one from the XAR classifier.

An additional mitigation strategy could be isolating and removing windows corresponding to transitions between activities. Indeed, as usually happens in HAR systems, a significant portion of the misclassifications occur while the subject transitions from one activity to another. During these transitions, the sensor events do not reflect any of the main target activities. Since the datasets we considered in this work do not have a label for transitions (e.g., ``\textit{Other}'' or ``\textit{Idle} classes) the classifier is always forced to choose among one of the target activities, many of the misclassifications occur for this reason. 
As a consequence, the most important semantic states described in the explanations of these misclassifications are inconsistent with the predicted activity.
While this phenomenon occurs for any XAR system,
its impact may be even more negative for LLM-based approaches, where the model aims at creating convincing explanations for semantic states that are, however, inconsistent with the predicted activity. Hence, being able to isolate transitions could, in addition to improving classification performance, further help in reducing the number of misleading explanations. 

Finally, a further strategy is using an agent-based approach. In this case, an LLM agent would be tasked with verifying that each rationale in the explanation is supported by the data and possibly providing an estimate of the risk of over-reliance.






\subsection{Hallucinations in LLMs}

Hallucinations in LLMs refer to instances where the model generates plausible but incorrect or fabricated information~\cite{huang2023survey}. 
In our domain, hallucinations can occur in activity classification and explanation generation.

Considering classification, hallucinations may lead to an incorrect activity, similar to mispredictions of traditional data-driven models. Additionally, hallucinations can manifest as outputs that do not comply with predefined instructions, such as predicting an activity outside the given set or producing results in an incorrect format. However, this second scenario never occurred in our experiments.

Considering explanations, hallucinations may introduce speculative or inconsistent reasoning. For instance, an LLM might invent correlations between sensor data to justify the predicted activity, creating an explanation that is linguistically coherent but not factually accurate. 
These cases can exacerbate over-reliance issues discussed in Section \ref{sec:over-reliance}.
In theory, LLMs may also generate entirely fabricated explanations poorly linked to the input, but we did not observe such extreme cases in our experiments. Nonetheless, the potential risk of hallucinations remains a challenge.

\subsection{Limitations in setups dominated by PIR  sensors}
\label{subsec:applicability}

Our experiments show that LLMs encode knowledge about ADLs. However, there are scenarios in which \zeroshot{} would fail. Indeed, we believe that \zeroshot{} can lead to good results only when the sensing infrastructure can properly capture the interaction of the subject with the environment.
Such sensing infrastructures are commonly considered for healthcare monitoring applications~\cite{grammatikopoulou2024assessing} relying on cost-effective off-the-shelf devices.

\begin{figure}[h!]
    \centering
    \includegraphics[width=0.8\linewidth]{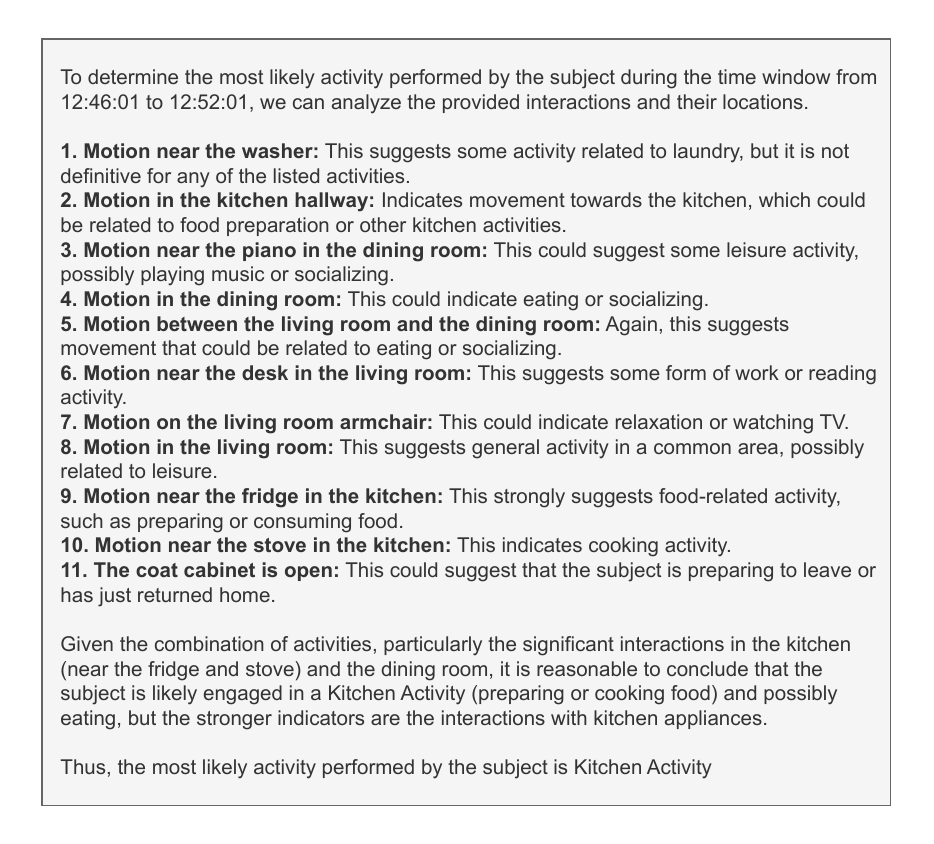}
    \caption{The Chain of Thought reasoning process of \zeroshot{} on the CASAS Milan dataset. The ground truth activity is \textit{work}.}
    \label{fig:casas}
\end{figure}

On the other hand, considering datasets like the well-known CASAS~\cite{cook2012casas}, most of the sensors used in the homes are PIR sensors only detecting the movements of the inhabitants with all the limits that these sensors are known to have. Even though supervised data-driven models can capture motion patterns driven by ground truth (with the negative consequence of struggling in generalizing across different environments), motion sensors alone can not precisely reveal the actual activity performed by the subject.
For this reason, LLMs struggle in mapping windows of motion sensor data into activities since it is challenging to use common-sense knowledge to determine ADLs from this data.
Figure~\ref{fig:casas} shows an example of the step-by-step ``reasoning'' (through CoT) done by \zeroshot{} on a window extracted from the CASAS Milan dataset. 

We adopted the segmentation window suggested in~\cite{arrotta2022dexar} (i.e., 360 seconds) and we used the house floor plan to determine the semantic states generated by the deployed sensors.
While the ground truth of this window is \textit{work}, the LLM predicted \textit{kitchen activity} since it considered, among the many fired motion sensors, the ones triggered in the kitchen as more important.

Based on the above considerations, it might not be worth spending money and resources to use LLMs when considering smart home deployments mostly based on PIR sensors. 

\subsection{Privacy issues, financial cost, and scalability}

In our experimental evaluation, we used a third-party cloud-based LLM that is known to perform very well on many tasks beyond ADL recognition. However, the adoption of these models in real deployments may have three drawbacks: privacy issues, financial cost, and scalability.

Considering privacy, personal data about human activities is sensitive, as it can reveal sensitive personal habits and medical conditions.
Hence, all data acquired in the smart home should be transmitted, processed, and stored with state-of-the-art security methods~\cite{bettini2025personal}. In particular, outsourcing such information to untrusted third-party providers (e.g., LLM services) exposes users to potential privacy issues.
Deploying an open-weights LLM on a private machine of a trusted entity in a trusted domain is a promising alternative to mitigate privacy problems.  
For instance, considering medical applications, dedicated servers could be deployed in a telemedicine infrastructure. It is also possible that, in the future, more lightweight language models %
(e.g., small models like Phi-4~\cite{abdin2024phi} or quantized versions of open-source models)
will potentially run on smart home gateways with an acceptable trade-off between performance and accuracy.

However, to reduce privacy risks using open-weights models, a careful configuration of security settings is required. This requires, for example, encrypting data on-transit and at-rest. From the security point of view, poorly configured servers may also increase the risks of unauthorized access and/or model replacements with tampered or backdoored versions that can behave maliciously. Moreover, LLMs can be tricked into executing malicious commands or generating harmful outputs if not properly sandboxed.

Considering financial cost, both \zeroshot{} and \llmxai{} require a remote call to the LLM APIs for each window. At the time we performed our experiments using the cloud-based gpt-4o model, a single window for \zeroshot{} was costing on average $0.0085\$$. Considering windows of $16$ seconds with $0.8$ of overlap, as we did for the MARBLE dataset, operating the system continuously at this price would cost approximately $230 \$$ per day, which is a very high cost. Despite the price for LLM usage is likely to decrease significantly with the design of more efficient and powerful models, this is an issue to consider.
On the other hand, even local deployments of open-weights models incur financial costs. Indeed, the most accurate open-weights LLMs currently still require powerful servers, resulting in costs not only for the hardware, but also for energy consumption. In this scenario, a telemedicine platform could optimize the costs by performing ADL recognition for a large number of subjects.

Finally, considering scalability, using third-party models introduces some bottlenecks to take into account, especially considering future smart homes that will likely be equipped with a significantly high number of sensing devices generating large inputs for the LLM. Indeed, third-party services usually have usage limits, for example in the number of requests, the number of tokens processed per minute, and the number of tokens for each request. To maximize throughput, implementing parallel API calls also requires adopting strategies like retries with exponential backoff, concurrency control, and request queuing.
Another factor impacting scalability is the latency per request, that is affected by network communication delays. 
On the other hand, to achieve the necessary scalability with open-weights models, the telemedicine platform should run multiple instances of the LLM across multiple machines with GPUs. This setup, with the adequate hardware resources, is crucial for potentially handling the many concurrent requests coming from numerous subjects, while also ensuring high availability and fault tolerance in case of individual instance failures. Considering future smart homes equipped with numerous sensing devices, we believe that recent open-weight models already allow processing inputs that are sufficiently large (e.g., \textit{Gemma 3 27b} has a context window of $128k$ tokens). As an alternative, distributing lightweight language models across the homes directly in edge devices (e.g., in the gateway) would further improve scalability. Hence, while open-weight models represent a promising direction, their deployment should be properly designed to take costs, scalability, and security configurations into account.

\subsection{Generalization and Personalization of \zeroshot}

Our \zeroshot\ approach leverages LLMs pretrained on vast and diverse textual corpora to perform zero-shot activity recognition without requiring labeled data. As demonstrated in our experiments across multiple datasets, \zeroshot{} achieves high recognition rates in home environments that differ in layout, object configurations, and user behavior.

However, the generalization capabilities of \zeroshot{} may still be limited considering individualized behavioral patterns, culturally specific routines, or unique home configurations not captured by the general LLM knowledge.

To mitigate this challenge for the smart home environments, \zeroshot{} system prompt requires details about the home infrastructure and the set of monitored objects (see Figure~\ref{fig:llme2e_system_prompt}). These cues guide the LLM in interpreting sensor data for ADLs recognition. As future work, we plan to expand the system prompt to incorporate user-specific context—such as habitual routines, social or cultural norms, and personal schedules to further personalize ADLs recognition.

While prompt engineering offers a lightweight method for adaptation, it may not suffice in all cases. Fine-tuning the LLM on labeled data from the target environment could, in principle, yield better personalization with individual users. However, such an approach is often constrained by the high costs, privacy risks, and logistical complexity of acquiring labeled data in real-world smart home deployments.

To address these limitations, semi-supervised learning strategies present a promising avenue. Techniques that combine unsupervised clustering with active learning (e.g., \cite{hiremath2022bootstrapping,arrotta2023selfact}) could enable minimal supervision by selecting only the most informative examples for labeling from the massive amount of unlabeled sensor data.

However, it is important to note that \zeroshot{} goes beyond classification; it also generates natural language explanations, which are the core focus of this paper. These explanations do not have ground-truth supervision and depend on the LLM's generative capabilities. Fine-tuning only the ADLs recognition task may risk degrading explanation quality. To the best of our knowledge, no fine-tuning method currently guarantees preservation of generative reasoning alongside predictive accuracy.

As an alternative, few-shot prompting offers a promising strategy for personalization. As also demonstrated in ADL-LLM~\cite{civitarese2024large}, by including a small number of labeled examples directly into the prompt, the model can be guided to improve both prediction and explanation without altering its parameters. This approach maintains the integrity of the generative capabilities while enabling user-specific adaptation.

Hence, we will explore hybrid personalization strategies integrating few-shot prompting with semi-supervised learning techniques, enabling \zeroshot{} to adapt more effectively to individual users.

This future line of research also opens up opportunities for long-term adaptation. Indeed, by continuously observing unlabeled data, it becomes possible to detect shifts in an individual's activity patterns over time. Integrating concept drift detection with active learning would allow the system to selectively update the labeled examples used in few-shot prompting, ensuring that the model remains aligned with the subject’s evolving routines and lifestyle.

\section{Conclusion and Future Work}

%


This paper explored how to leverage LLMs for XAR, showing the possible benefits, but also evaluating the drawbacks and risks.
%
Our \zeroshot{} method is particularly promising for real-world applications, where labeled data collection is not feasible. Indeed,  its ability to recognize activities in a zero-shot setting (i.e. when no training data is available) combines with the generation of explanations considered as high quality according to our user surveys. 
The \llmxai{} method demonstrates how LLMs can potentially generate explanations for any XAR method, thus improving their understandability for non-expert users.

Our critical evaluation highlighted the main problems in generating explanations with LLMs and illustrated several mitigation strategies that may be taken. Overall, we believe that once these challenges are addressed, the benefits will greatly outweigh the drawbacks and LLMs will become a powerful tool for implementing XAR systems in the wild. 



As part of a collaboration project with neurologists~\footnote{https://ecare.unimi.it/pilots/serenade/}, we are collecting a large unlabeled dataset from smart homes of individuals with Mild Cognitive Impairment (MCI). Hence, in future work, we plan to test our methods on this dataset, possibly also exploring open weights Large and Small Language Models.

 



\begin{acks}
This work was supported in part by projects MUSA and FAIR under the NRRP MUR program funded by the EU-NGEU. Views and opinions expressed are however those of the authors only and do not necessarily reflect those of the European Union or the Italian MUR. Neither the European Union nor the Italian MUR can be held responsible for them.
\end{acks}

\bibliographystyle{ACM-Reference-Format}
\bibliography{references}
\end{document}